\def\year{2021}\relax
\documentclass[letterpaper]{article} 
\usepackage{aaai21}  
\usepackage{times}  
\usepackage{booktabs}
\usepackage{graphicx}
\usepackage{multirow}
\usepackage{algorithm}
\usepackage{color}
\usepackage{xcolor}
\usepackage{subcaption}
\usepackage{colortbl}
\usepackage{microtype}
\usepackage{algpseudocode}
\usepackage{algorithmicx}
\usepackage{multirow}
\newcommand{\tabincell}[2]{\begin{tabular}{@{}#1@{}}#2\end{tabular}}  

\usepackage{graphicx}
\usepackage{caption}
\usepackage{soul}
\usepackage{ulem}
\usepackage{xcolor}
\definecolor{yellow}{RGB}{255,255,0}
\definecolor{green}{RGB}{34,139,34}

\usepackage{amsfonts,amssymb}
\usepackage[namelimits]{amsmath}

\usepackage{helvet} 
\usepackage{courier}  
\usepackage[hyphens]{url}  
\usepackage{graphicx} 
\usepackage{natbib}  
\usepackage{caption} 
\begin{document}
\title{Iterative Utterance Segmentation for Neural Semantic Parsing }

\author{
    Yinuo Guo\textsuperscript{\rm 1}\thanks{Work done during an internship at Microsoft Research Asia.}, Zeqi Lin\textsuperscript{\rm 2}, Jian-Guang Lou\textsuperscript{\rm 2}, Dongmei Zhang\textsuperscript{\rm 2} \\
}
\affiliations{
    \textsuperscript{\rm 1}Key Laboratory of Computational Linguistics, \\
    School of EECS, Peking University,\\
    \textsuperscript{\rm 2}Microsoft Research Asia \\
    \textsuperscript{\rm 1}gyn0806@pku.edu.cn,\\
    \textsuperscript{\rm 2}\{Zeqi.Lin, jlou, dongmeiz\}@microsoft.com
}

\maketitle

\begin{abstract}
Neural semantic parsers usually fail to parse long and complex utterances into correct meaning representations, due to the lack of exploiting the principle of compositionality. To address this issue, we present a novel framework for boosting neural semantic parsers via iterative utterance segmentation.
Given an input utterance, our framework iterates between two neural modules: a \textit{segmenter} for segmenting a span from the utterance, and a \textit{parser} for mapping the span into a partial meaning representation. Then, these intermediate parsing results are composed into the final meaning representation. One key advantage is that this framework does not require any handcraft templates or additional labeled data for utterance segmentation: we achieve this through proposing a novel training method, in which the parser provides pseudo supervision for the segmenter. Experiments on \textsc{Geo}, \textsc{ComplexWebQuestions} and \textsc{Formulas} show that our framework can consistently improve performances of neural semantic parsers in different domains.
On data splits that require compositional generalization, our framework brings significant accuracy gains: \textsc{Geo} $63.1\to 81.2$, \textsc{Formulas} $59.7\to 72.7$, \textsc{ComplexWebQuestions} $27.1\to 56.3$.
\end{abstract}

\section{Introduction}
Semantic parsing is the task of mapping natural language utterances to machine interpretable meaning representations.
Many semantic parsing methods are based on the principle of semantic compositionality (aka, compositional semantics)~\citep{Pelletier1994}, of which the main idea is to put together the meanings of utterances by combining the meanings of the parts~\citep{Zelle1996,zettlemoyer-collins-2005-learning, zettlemoyer-collins-2007-online, liang-etal-2011-learning, berant-etal-2013-semantic, pasupat-liang-2015-compositional, berant-liang-2015-imitation}.
However, these methods suffer from heavy dependence on handcrafted grammars, lexicons, and features.

To overcome this problem, many neural semantic parsers have been proposed and achieved promising results~\citep{jia-liang-2016-data, dong-lapata-2016-language, ling-etal-2016-latent, dong-lapata-2018-coarse, shaw-etal-2019-generating}.
However, due to the lack of capturing compositional structures in utterances, neural semantic parsers usually have poor generalization ability to handle unseen compositions of semantics~\citep{finegan-dollak-etal-2018-improving}.
For example, a parser trained on ``\emph{How many rivers run through oklahoma?}'' and ``\emph{Show me states bordering colorado?}'' may not perform well on ``\emph{How many rivers run through the states bordering colorado?}''.

\begin{table}
\small
\centering
\begin{tabular}{cp{6cm}}
\hline
\textbf{Q} & \textcolor{red}{How many} \textcolor{green}{rivers run through} \textcolor{blue}{the states bordering colorado}\\
\textbf{M} & $count(river(traverse\_2(state($\\
& $next\_to\_2(stateid('colorado'))))))$\\
\hline
\textbf{\textcolor{blue}{Q1}} & \textcolor{blue}{the states bordering colorado}\\
\textbf{M1} & $state(next\_to\_2(stateid('colorado')))$\\
\hline
\textbf{\textcolor{green}{Q2}} & \textcolor{green}{rivers run through} \textcolor{blue}{$\$state\$$}\\
\textbf{M2} & $river(traverse\_2(\$state\$))$\\
\hline
\textbf{\textcolor{red}{Q3}} & \textcolor{red}{How many} \textcolor{green}{$\$river\$$}\\
\textbf{M3} & $count(\$river\$)$\\
\hline
\end{tabular}
\caption{An example of iterative utterance segmentation for semantic parsing.
The first cell shows an utterance ($Q$) and its meaning representation ($M$).
The following three cells show how we iterate between:
(1) segmenting the utterance to obtain a simpler span ($Q1$/$Q2$/$Q3$);
(2) parsing the span into a partial meaning representation ($M1$/$M2$/$M3$).
Finally, we compose $M1$, $M2$ and $M3$ into the final result ($M$).}
\label{table:example}
\end{table}

\begin{table*}
\small
\centering
\begin{tabular}{cp{12cm}}
\hline
\textbf{Dataset} & \textbf{Example} \\
\hline
\multirow{2}{*}{\textsc{Geo}} & \emph{x: ``How many rivers run through the states bordering colorado?''}\\
& \emph{y: count(river(traverse\_2(state(next\_to\_2(stateid('colorado'))))))} \\
\hline
\multirow{6}{*}{\tabincell{c}{\textsc{Complex}\\\textsc{WebQuestions}}} & \emph{x: ``What is the mascot of the school where Thomas R. Ford is a grad student?''}\\
& \emph{y: SELECT ?x WHERE \{}\\
& \quad\quad\emph{?c ns:education.educational\_institution.students\_graduates ?k .}\\
& \quad\quad\emph{?k ns:education.education.student ns:m.0\_thgpt .}\\
& \quad\quad\emph{?c ns:education.educational\_institution.mascot ?x \}}\\
\hline
\multirow{2}{*}{\tabincell{c}{\textsc{Formulas}}} & \emph{x: ``What is the smaller value between F14 divided by E14 and the largest number in A1:D10.''}\\
& \emph{y: MIN(F14/E14, MAX(A1:D10))}\\
\hline
\end{tabular}
\caption{Examples of natural language utterances ($x$) paired with their structured meaning representations ($y$) from our experimental datasets.}
\label{table:pairs}
\end{table*}

In this paper, we propose a novel framework to boost neural semantic parsers with the principle of compositionality~\citep{Pelletier1994}.
It iterates between segmenting a span from the utterance and parsing it into a partial meaning representation.
Table \ref{table:example} shows an example.
Given an utterance ``\emph{How many rivers run through the states bordering colorado?}'', we parse it through three iterations:
(1) we segment a span ``\emph{the states bordering colorado}'' from the utterance, and parse it into $state(next\_to\_2(stateid('colorado')))$;
(2) as the utterance is reduced to ``\emph{How many rivers run through \$state\$}?'', we segment a span ``\emph{rivers run through \$state\$}'' from it, and parse it into $river(traverse\_2(\$state\$))$;
(3) the utterance is further reduced to ``\emph{How many \$river\$?}'', and we parse it into $count(\$river\$)$.
We compose these partial meaning representations into the final result.

Our framework consists of two neural modules: an utterance segmentation model (\emph{segmenter} for short) and a base parser (\emph{parser} for short).
The former is in charge of segmenting a span from an utterance, and the latter is in charge of parsing the span into its meaning representation.
These two modules work together to parse complex input utterances in a divide-and-conquer fashion.

One key advantage of this framework is that it does not require any handcraft templates or additional labeled data for utterance segmentation: we achieve this through proposing a novel training method, in which the base parser provides pseudo supervision to the utterance segmentation model.
Specifically:
we train a preliminary base parser on the original train data; then, for each train sample $(x, y)$, we leverage the preliminary base parser to collect all reasonable spans~(those can be parsed to a part of $y$ by the preliminary base parser).
Finally, we use these collected spans as pseudo supervision signals for training the utterance segmentation model, without requiring any handcraft templates or additional labeled data.




In summary, our proposed framework has four advantages:
(1) the base parser learns to parse simpler spans instead of whole complex utterances, thus alleviating the training difficulties and improving the compositional generalization ability;
(2) our framework is flexible to incorporate various popular encoder-decoder models as the base parser;
(3) our framework does not require any handcraft templates or additional labeled data for utterance segmentation;
(4) our framework improves the interpretability of neural semantic parsing by providing explicit alignment between spans and partial meaning representations.

We conduct experiments on three datasets: \textsc{Geo}~\citep{Zelle1996}, \textsc{ComplexWebQuestions}~\citep{talmor-berant-2018-web}, and \textsc{Formulas} (a new dataset introduced in this paper).
They use different forms of meaning representations: FunQL, SPARQL, and Spreadsheet Formula. Experimental results show that our framework consistently improves the performances of neural semantic parsers in different domains.
On data splits that require compositional generalization, our framework brings significant accuracy gain: \textsc{Geo} $63.1\to 81.2$, \textsc{Formulas} $59.7\to 72.7$, \textsc{ComplexWebQuestions} $27.1\to 56.3$.

\begin{figure*}[htbp]
    \small
	\centering
	\includegraphics[width = 0.85\textwidth]{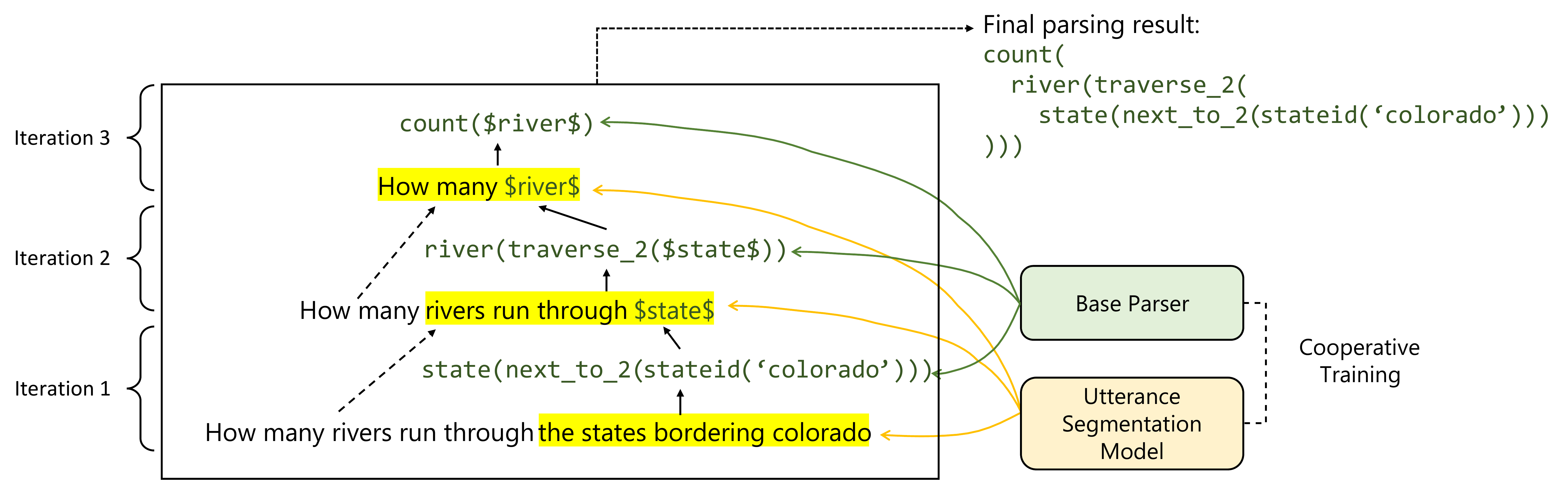}
	\caption{Framework overview. Given a natural language utterance, the framework iteratively segments \colorbox{yellow}{a span} from it and parses the span to \emph{\textcolor{green}{a partial meaning representation}}. We obtain span meanings via neural semantic parsing, then piece them to assemble whole-utterance meaning.}\label{fig:example}
\end{figure*}

\section{Related Work}
\subsection{Semantic Parsing}

There are two major paradigms of semantic parsing:
compositional semantic parsing~\citep{Zelle1996,zettlemoyer-collins-2005-learning, zettlemoyer-collins-2007-online, liang-etal-2011-learning, berant-etal-2013-semantic, pasupat-liang-2015-compositional, berant-liang-2015-imitation},
and neural semantic parsing~\citep{jia-liang-2016-data, dong-lapata-2016-language, ling-etal-2016-latent, dong-lapata-2018-coarse, shaw-etal-2019-generating}.
Our work aims to combine their respective advantages.

In neural semantic parsing, various efforts have been made to leverage the syntax of meaning representations (typically, the tree structures) to enhance decoders~\citep{dong-lapata-2016-language, xiao-etal-2016-sequence, krishnamurthy-etal-2017-neural, cheng-etal-2017-learning, yin-neubig-2017-syntactic, rabinovich-etal-2017-abstract, guo-etal-2019-towards}.
In these works, encoders treat input utterances as sequential tokens, without considering their compositional structures. On the other hand, some researchers focus on exploring data augmentation techniques to provide a compositional inductive bias in models~\cite{jia-liang-2016-data,andreas2019good}.
However, they rely on exact matching of spans, which work well on word/phrase-level re-combination or simple domains (e.g., \textsc{GEO} and \textsc{SCAN}~\cite{lake2018generalization}), but is not suitable for more complex scenarios (e.g., diverse subsentences in \textsc{ComplexWebQuestions}). Therefore, the lack of compositional generalization ability is still a challenging problem in neural semantic parsing~\citep{finegan-dollak-etal-2018-improving, keysers2020measuring}. 

\subsection{Utterance Segmentation}

In Question Answering, question segmentation has been successfully applied to help answer questions requiring multi-hop reasoning~\citep{kalyanpur2012fact, talmor-berant-2018-web, min-etal-2019-multi, qi-etal-2019-answering}.
A key challenge in these works is to derive supervision for question segmentation.
\citet{kalyanpur2012fact} segments questions based on predominantly lexico-syntactic features.
\citet{talmor-berant-2018-web} leverages simple questions to derive distant supervision.
\citet{min-etal-2019-multi} uses additional labeled data to fine-tune a BERT-based model;
\citet{qi-etal-2019-answering} utilizes the longest common strings/sequences between utterances and their supporting context documents for segmentation.

In Semantic Parsing, \citet{zhang-etal-2019-complex} proposes HSP, a novel hierarchical semantic parsing method, which utilizes the decompositionality of complex utterances for semantic parsing.
This method requires \emph{(utterance, sub-utterances, meaning representation)} instances for training.
\citet{pasupat2019span} proposes a span-based hierarchical semantic parsing method for task-oriented dialog.
This method requires span-based annotations for training.

\section{Framework}

\subsection{Problem Statement}
Our goal is to learn semantic parsers from $\mathcal{D}$, a set of instances of natural language utterances  paired with their structured meaning representations.
We wish to estimate $p(y|x)$, the conditional probability of meaning representation $y$ given utterance $x$.
Table \ref{table:pairs} shows examples from different datasets.

\subsection{Framework Overview}
\label{section:model_overview}
Neural semantic parsers usually have poor generalization ability to handle unseen compositions of semantics.
To address this problem, we propose to segment a complex utterance into simpler sub-utterances.
We obtain partial meanings via neural semantic parsing, and then compose them together as the entire meaning representation.

Specifically, our framework consists of an utterance segmentation model and a base neural semantic parser.
It parses utterances in an iterative ``segment-and-parse'' fashion. The $k$-th iteration process is detailed in the following four steps:

\begin{enumerate}
    \item \textbf{Segmentation.} Given an input utterance $x^{(k)}$ ($x^{(1)}$ is the original input utterance), our utterance segmentation model predicts a span $\hat{x}=x^{(k)}_{i:j} (1\leq i < j \leq |x^{(k)}|)$, representing an independent clause of $x^{(k)}$.
    \begin{equation}\nonumber
    \footnotesize
    \underbrace{\text{How many rivers run through}\overbrace{\text{ the states bordering colorado}}^{\mbox{$\hat{x}$}}\text{ ?}}_{\mbox{$x^{(k)}$}}
    \end{equation}
    \item \textbf{Parsing.} The base parser maps span $\hat{x}$ to a partial meaning representation $\hat{y}$.
    \begin{equation}\nonumber
    \hat{x}\to \overbrace{state(next\_to\_2(stateid('colorado'))}^{\mbox{$\hat{y}$}}
    \end{equation}
    \item \textbf{Reducing.}
    Based on $z$, we reduce $x^{(k)}$ to $x^{(k+1)}$ through substituting $\hat{x}$ with $d(\hat{y})$:
    \begin{equation}
        x^{(k+1)}=[\hat{x}\mapsto d(\hat{y})]x^{(k)}
    \end{equation}
    where the notation $[u\mapsto v]p$ refers to the result of replacing span $u$ in $p$ (an utterance or meaning representation) with $v$, and $d(\hat{y})$ is the denotation type (i.e., answer type) of $\hat{y}$\footnote{Implementation details of $d(\hat{y})$ in different domains are presented in Section \ref{section:implementation}.}.
    \begin{equation}\nonumber
    \footnotesize
    \underbrace{\text{How many rivers run through}\overbrace{\text{ \$state\$}}^{\mbox{$d(\hat{y})$}}\text{ ?}}_{\mbox{$x^{(k+1)}$}}
    \end{equation}
    \item \textbf{Iteration.} If $s\neq x^{(k)}$, start the $(k+1)$ -th iteration.
    Otherwise, stop the iteration process, and compose the partial meaning representations produced in the iterations into the final result.
    \begin{equation}\nonumber
    \footnotesize
    \underbrace{\text{How many}\overbrace{\text{ rivers run through \$state\$}}^{\mbox{$\hat{x}$}}\text{ ?}}_{\mbox{$x^{(k+1)}$}}
    \end{equation}
\end{enumerate}

Figure \ref{fig:example} illustrates how our framework parses a natural language utterance to its meaning representation through iterative utterance segmentation.

\subsection{Probabilistic Reformulation}

To be more formal, in this section we rephrase Section \ref{section:model_overview} using conditional probability.

Firstly, the conditional probability $p(y|x)$ is decomposed into a two-stage generation process:

\begin{equation}\label{eq:p0}
    p(y|x) = \sum_{\hat{x}\subset x} p(y|x,\hat{x})p(\hat{x}|x)
\end{equation}
where $p(\hat{x}|x)$ represents the conditional probability of segmenting a span $\hat{x}$ from $x$, and $p(y|x,\hat{x})$ represents the conditional probability of parsing $x$ to meaning representation $y$ after the segmentation.

Then, we further decompose $p(y|x,\hat{x})$ into a two-stage generation process:

\begin{equation}\label{eq:p1}
    p(y|x,\hat{x}) = \sum_{\hat{y}} p(y|x,\hat{x},\hat{y})p(\hat{y}|x,\hat{x})
\end{equation}

As $\hat{x}$ represents an independent clause in $x$ (which means that $\hat{y}$ is independent of $x$ given $\hat{x}$), we have $p(\hat{y}|x,\hat{x})=p(\hat{y}|\hat{x})$.

Now we consider $p(y|x,\hat{x},\hat{y})$:


\begin{equation}\label{eq:p2}
\begin{split}
    p(y|x,\hat{x},\hat{y}) & = p(reduced\_y|reduced\_x) \\
    reduced\_x & = [\hat{x} \mapsto d(\hat{y})]x \\
    reduced\_y & = [\hat{y} \mapsto d(\hat{y})]y \\
\end{split}
\end{equation}
This corresponds to the iteration mechanism in our framework.

According to Equation \ref{eq:p0}, \ref{eq:p1} and \ref{eq:p2}, the overall conditional probability $p(y|x)$ 
consists of three components:

\begin{itemize}
    \item $p(\hat{x}|x)$: an utterance segmentation model that predicts a span $\hat{x}$ from $x$.
    Section \ref{section:segmentation} details this component.
    \item $p(\hat{y}|\hat{x})$: a base parser that map $\hat{x}$ to a partial meaning representation $\hat{y}$.
    In this work, we make it a neural semantic parser based on encoder-decoder architecture.
    A simple implementation is detailed in Section \ref{section:base_parser}.
    \item $p(y|x,\hat{x},\hat{y})$: the iteration mechanism.
\end{itemize}

\subsection{Base Parser}
\label{section:base_parser}
Our framework is flexible to incorporate various popular encoder-decoder models as the base parser.
Without loss of generality, we use a typical sequence-to-sequence semantic parsing model proposed by \citet{dong-lapata-2016-language} as a default base parser.

The encoder is a bi-directional RNN with gated recurrent units (GRU, \citet{cho2014learning}):

\begin{equation}
    \mathbf{enc} = \text{Bi-GRU}_{enc}(\mathbf{s})
\end{equation}

The decoder is another GRU network with attention component~\citep{luong-etal-2015-effective}:

\begin{equation}
    \mathbf{dec} = \text{Attn-Bi-GRU}_{dec}(\mathbf{enc})
\end{equation}

Then, $p(\hat{y}_t|\hat{y}_{<t},\hat{x})$, the conditional probability for generating the next word $\hat{y}_t$, is estimated via

\begin{equation}
    p(\hat{y}_t|\hat{y}_{<t},\hat{x}) = softmax_{z_t}(Linear(\mathbf{dec}_t))
\end{equation}

The base parser will be initially trained on $\mathcal{D}$ (the original train set), and further fine-tuned on pseudo supervision signals (detailed in Section \ref{section:train}).

\subsection{Utterance Segmentation}
\label{section:segmentation}

We train an utterance segmentation model $Seg$ that learns to predict a span (i.e., the start position $i$ and end position $j$ of the span) from $x$.

\subsubsection{Span Prediction}
In $Seg$, we use a GRU to encode $x$:

\begin{equation}
    \mathbf{U} = \text{GRU}_{seg}(x)\in\mathbb{R}^{m\times u}
\end{equation}

Then, the span $\hat{x}$ is predicted via:

\begin{equation}
\begin{split}
    p(i|x) & = softmax_i(\mathbf{U}\mathbf{W}_I)\\
    p(j|x) & = softmax_j(\mathbf{U}\mathbf{W}_J)
\end{split}
\end{equation}
where $\mathbf{W}_I, \mathbf{W}_J\in \mathbb{R}^{u}$.

\subsection{Training}
\label{section:train}

\subsubsection{Pseudo Supervision}
Training the utterance segmentation model is non-trivial, because: (1) there is usually no labeled data of how utterances should be segmented; (2) 
we do not manage to use handcraft utterance segmentation templates aiming to have better generalization ability.
To address this problem, we propose to leverage the base parser to derive pseudo supervision for utterance segmentation.

Take the aforementioned utterance ``\emph{How many rivers run through the states bordering colorado?}'' as an example.
As described in \ref{section:base_parser}, we have a preliminary base parser which is trained on the original train set $\mathcal{D}$.
For each span $\hat{x}$ in utterance $x$, we use this preliminary base parser to check whether this span is a ``good'' span.
Specifically: we use the preliminary base parser to parse $\hat{x}$ into $\hat{y}$; if $\hat{y}$ is a part of the target meaning representation $y$, we call $\hat{x}$ a good span.
For example, ``\emph{the states bordering colorado}'' is a good span:
\begin{equation}\nonumber
    \footnotesize
    \text{How many rivers run through}\overbrace{\text{ the states bordering colorado}}^{\color{green}{state(next\_to\_2(stateid('CO'))}}\text{ ?}
\end{equation}

As a comparison, ``\emph{run through the states bordering colorado}'' is not a good span (because the corresponding $\hat{y}$ is not a part of $y$):
\begin{equation}\nonumber
    \footnotesize
    \text{How many rivers}\overbrace{\text{ run through the states bordering colorado}}^{\color{red}{state(traverse\_2(stateid('CO'))}}\text{ ?}
\end{equation}

Each utterance may have several good spans.
We define the best span as the shortest good span, with a constraint that the parsing result of the best span should not contain any ``ghost'' entity.
For example, ``\emph{the states bordering}'' is regarded as a good span, since the preliminary base parser parses it into ``$state(traverse\_2(stateid('colorado')))$''.
We think ``$stateid('colorado')$'' is a ghost entity, as it is not mentioned in the span.
We do not think such spans should be segmented, so we restrict that none of they should be the best span.
If a span has no best span, we regard itself as its best span.

For each utterance in the train set $\mathcal{D}$, we use the best span as a pseudo supervision signal for training the utterance segmentation model.
We denote all these pseudo supervision signals as $\mathcal{A}$.




\subsubsection{Training Objective}

The overall training objective is:
\begin{equation}
\begin{split}
    \mathcal{J}(\phi,\theta) & = \mathcal{J}_{Seg}(\phi) + \mathcal{J}_{Bp}(\theta)\\
    \mathcal{J}_{Seg}(\phi) & = \sum_{(x,\hat{x})\in\mathcal{A}}\log{p(\hat{x}|x)}\\
    \mathcal{J}_{Bp}(\theta) & = \sum_{(x,y)\in\mathcal{D}}\log{p(y|x)} + \sum_{(\hat{x},\hat{y})\in\mathcal{\hat{D}}}\log{p(\hat{y}|\hat{x})}
\end{split}
\end{equation}
where $\mathcal{J}_{Seg}(\phi)$ is the training objective of the utterance segmentation model, and $\mathcal{J}_{Bp}(\theta)$ is the training objective of the base parser.
$\phi$ and $\theta$ refer to learnable parameters in them respectively.
$\hat{\mathcal{D}}$ consists of two parts: (1) best spans paired with their partial meaning representations; (2) reduced utterances paired with their partial meaning representations.


\subsection{Inference}
At inference time, we iteratively ``segment-and-parse'' the utterance.
In the $k$-th iteration, we predict a span $\hat{x}^*$ from $x^{(k)}$ by $\hat{x}^*=\arg\max_{\hat{x}}{p(\hat{x}|x^{(k)})}$, then parse $\hat{x}^*$ to $\hat{y}^*$ by $\hat{y}^*=\arg\max_{\hat{y}}{p(\hat{y}|\hat{x}^*)}$.
Then, we let $x^{(k+1)}=[\hat{x}^*\mapsto d(\hat{y}^*)]x^{(k)}$ and start the $(k+1)$ -th iteration, until $x^{(k+1)}=x^{(k)}$.
Partial meaning representation outputs ($\hat{y}^*$, ...) during these iterations are composed deterministically to form the final meaning representation.

\section{Experiments}

\subsection{Datasets}
\label{section:datasets}
We conduct experiments on three datasets: \textsc{Geo}~\citep{Zelle1996}, \textsc{ComplexWebQuestions}~\citep{talmor-berant-2018-web} and \textsc{Formulas} (a new dataset). 
Examples of \emph{(utterance, meaning representation)} instances in these three datasets are shown in Table \ref{table:pairs}.

\subsubsection{\textsc{Geo}}

GEO is a standard semantic parsing benchmark about U.S. geography~\citep{Zelle1996}, which consists of 880 English questions paired with their meaning representations.
These meaning representations can be expressed as four equivalent expressions, namely $\lambda-$calculus, Prolog, SQL, and FunQL~(Functional Query Language)~\citep{kate2005learning}. Considering the compatibility with the proposed framework, we use FunQL in this paper.

GEO can be splitted into train/test sets in two different ways:
(1) \emph{Standard split} \citep{zettlemoyer-collins-2005-learning}: this split ensures that no natural language question is repeated between the train and test sets;
(2) \emph{Compositional split} \citep{finegan-dollak-etal-2018-improving}: this split ensures that neither questions nor meaning representations(anonymizing named entities) are repeated .
We evaluate our framework on both two splits: standard split for comparing with previous systems, and compositional split for measuring the compositional generalization ability.
To have a fair comparsion with previous work, we use the preprocessed dataset provided by \citet{dong-lapata-2016-language}, which does lemmatization for each natural language question and replaces entity mentions by numbered markers.
For example, ``\emph{How many rivers run through the states bordering colorado}'' will be preproccessed to ``\emph{How many river run through the state border state\_0}''.


\subsubsection{\textsc{ComplexWebQuestions}}
This dataset~\citep{talmor-berant-2018-web} contains questions paired with SPARQL queries for Freebase~\citep{bollacker2008freebase}.
``Complex'' means that all questions in this dataset are long and complex, requiring multi-hop reasoning to solve.
There are 27,734/3,480/3,475 train/dev/test examples in this dataset.
Following the settings in \citet{zhang-etal-2019-complex}, we use the \emph{V1.0} version of this dataset, and replace entities in SPARQL queries with placeholders during training and inference.
In the test set, 11.9\% SPARQL queries (after anonymizing entities, numbers and dates) have never been seen in the train set.
We use this subset to measure the compositional generalization ability.
\subsubsection{\textsc{Formulas}}
People need to write formulas to manipulate/analyze their tabular spreadsheet data, but it is difficult for them to learn and remember the names and usages of various functions.
We want to help people interact with their tabular spreadsheet data using natural language:
input natural language commands (e.g., ``\emph{show me the largest number in A1:D10}''), and the corresponding formulas (e.g. \emph{MAX(A1:D10)}) will be returned automatically.
Therefore, we construct \textsc{Formulas}, a semantic parsing dataset containing \emph{(natural language command, spreadsheet formula)} instances.
We invited 16 graduate students majoring in computer science (all of them master spreadsheet formulas) as volunteers for manual annotations.
They annotate 1,336 formulas (407 single-function formulas and 929 compound formulas) for 30 real-world spreadsheet files. 
There are total 28 functions used in these formulas, e.g., \emph{SUM}, \emph{MAX}, \emph{FIND}, \emph{CONCAT} and \emph{LOOKUP}.
We randomly split this dataset into three parts: 800 instances for training, 268 instances for development, and 268 instances for test.
We also replace entity mentions by typed markers, for example, replacing ``\emph{A1:D10}'' by `` \emph{\$cellrange\$}''.
In the test set, 29.8\% formulas (after anonymizing entities, numbers and dates) have never been seen in the train set.
We use this subset to measure the compositional generalization ability.

\subsection{Implementation Details}
\label{section:implementation}
\subsubsection{Base Parsers}
Besides the Seq2Seq base parser introduced in Section \ref{section:base_parser}, we also implement (1) Seq2Tree~\cite{dong-lapata-2016-language} for \textsc{Geo} and \textsc{Formulas}; (2) Transformer~\cite{vaswani2017attention} for \textsc{ComplexWebQuestions}.
We do not utilize transformer on \textsc{GEO}/\textsc{Formulas}, because they only have hundreds of instances for training.
For \textsc{ComplexWebQuestion}, we switch to transformer in order to have a fair comparison with \textsc{HSP}~(state-of-the-art).

\subsubsection{Model Configuration}
We set the dimension of word embedding to 300.
In \textsc{Seq2Seq}/\textsc{Seq2Tree} base parsers, we set the dimension of hidden vector to 512.
In the utterance segmentation model, the dimension of hidden vector is set to 300.
We use the Adam optimizer with default settings (in \textsc{PyTorch}) and a dropout layer with the rate of 0.5. The training process lasts 100 epochs with batch size 64.
For the \textsc{Transformer} base parser in \textsc{ComplexWebQuestions}, we follow the settings in \citet{vaswani2017attention}.

\subsubsection{Restricted Copy Mechanism}
HSP~\citep{zhang-etal-2019-complex} incorporates copy mechanism~\citep{gu-etal-2016-incorporating} to tackle OOV tokens in \textsc{ComplexWebQuestions} and shows its high importance. Observing that most OOV tokens are values (e.g., numbers and dates), we further propose a restricted copy mechanism (denoted as \textbf{\textsc{Copy*}}): recognize values in utterances via regex-based patterns, and constrain that only these terms can be copied.
\subsubsection{Span Substitution}
In our framework, we need to implement $d(z)$ for each domain, which represents the denotation type of partial meaning representation $z$ (see Equation \ref{eq:p2}).
For \textsc{Geo} and \textsc{Formulas}, we directly infer $d(z)$ from $z$ through a syntax-directed translation algorithm~\cite{Aho:1969:SDT:1739930.1740037}:
For \textsc{Geo}, $d(z)$ can be ``\$state\$'', ``\$city\$'', ``\$river\$'', ``\$place\$'', ``\$mountain\$'' or ``\$lake\$'';
For \textsc{Formulas}, $d(z)$ can be ``\$number\$'', ``\$string\$'', ``\$date\$'', ``\$bool\$'', ``\$cell\$'' or ``\$cellrange\$''.
For \textsc{ComplexWebQuestions}, we define $d(z)$ as the first noun phrase in the span, and we compute $[u\mapsto v]y$ ($y$, $u$ and $v$ are meaning representations) based on SPARQL semantics (rather than the basic string matching-based definition in Section \ref{section:model_overview}).
Moreover, as \textsc{ComplexWebQuestions} contains not only nesting questions but also conjunctive questions, we use a simple but effective heuristic rule to extend our framework: if a span is at the beginning of the utterance, we combine generated partial SPARQL queries using conjunction operation; otherwise we combine them using nesting operation.

\subsubsection{Pre-Training}
In our framework, we need to pre-train the base parser to make it capable to parse simple spans at initial time.
For \textsc{Geo} and \textsc{Formulas}, we pre-train the base parser using the original training dataset $\mathcal{D}$ which contains many simple instances.
In contrast, all questions in \textsc{ComplexWebQuestions} are complex, thus the original training dataset is not suitable for pre-training the base parser.
\textsc{ComplexWebQuestions} is constructed through (1) simple instance combination, and (2) crowdsourcing rephrasing.
We denote these simple instances as $\mathcal{D}_{seed}$.
However, $\mathcal{D}_{seed}$ is also not suitable for pre-training the base parser, since questions in it are lack of diverse linguistic expression~(half of them are machine-generated via fixed templates).
To tackle this issue, we use a heuristic algorithm to rephrase questions in $\mathcal{D}_{seed}$.
Suppose that: $(x,y)\in\mathcal{D}_{seed}$, and there exists $(\tilde{x},\tilde{y})\in\mathcal{D}$ and $(x',y')\in\mathcal{D}_{seed}$ such that $\tilde{y}$ is a combination of $y$ and $y'$.
We find a span $s$ in $\tilde{x}$, which maximizes $score(s|x, \tilde{x}, x')$:
\begin{equation}
\label{eq:split}
\begin{split}
    score(s|x, \tilde{x}, x') & = sim(s, x) \\
               & + sim([s\mapsto d(y')]\tilde{x}, x') \\
    sim(s,s') & = \sum_{w\in s}\max_{w'\in s'}cos(\mathbf{w},\mathbf{w'}) \\
\end{split}
\end{equation}
where $\mathbf{w}$ and $\mathbf{w'}$ represent fasttext word embeddings~\citep{bojanowski2016enriching} of word $w$ and $w'$, respectively.
Then, we treat $s$ as a rephrase of $x$ and add $(s, y)$ to $\mathcal{D'}$.
We use $\mathcal{D'}$ to pre-train our base parser.
\begin{table}[t]
    \small
    \centering
    \begin{tabular}{lc}
        \hline
       Compositional Semantic Parser & Accuracy \\
        \hline
        \textsc{ZC07} \small{\citep{zettlemoyer-collins-2007-online}} & 86.1\%\\
        \textsc{DCS} \small{\citep{liang-etal-2011-learning}} & 87.9\%\\
        \textsc{TISP} \small{\citep{zhao-huang-2015-type}} & 88.9\%\\
        \hline
        Neural Semantic Parser & \\
        \hline
        \textsc{Seq2Seq} \small{\citep{dong-lapata-2016-language}} & 84.6\%\\
        \textsc{Seq2Tree} \small{\citep{dong-lapata-2016-language}} & 87.1\%\\
        \textsc{DataRecombine} \small{\citep{jia-liang-2016-data}} & 89.3\%\\
        \textsc{Scanner}$^*$ \small{\citep{cheng-etal-2017-learning}} & 86.7\%\\
        \textsc{ASN} \small{\citep{rabinovich-etal-2017-abstract}} & 87.1\%\\
        \textsc{Coarse2Fine} \small{\citep{dong-lapata-2018-coarse}} & 88.2\%\\
        \textsc{GNN}$^*$ \small{\citep{shaw-etal-2019-generating}} & 89.3\%\\
        \qquad+BERT$^*$ & 92.5\%\\
        \hline
        \textsc{Seq2Seq}$^*$ & 85.6\%\\
        \qquad \textbf{+ PDE}$^*$ & \textbf{90.7}\%\\
        \textsc{Seq2Tree}$^*$ & 84.2\%\\
        \qquad \textbf{+ PDE}$^*$ & 88.9\%\\
        \hline       
    \end{tabular}
    \caption{Accuracies on \textsc{Geo} (standard split). Methods marked with $*$ use FunQL as meaning representations.}
    \label{table:geo_result}
\end{table}
\begin{table}[t]
    \small
    \centering
    \begin{tabular}{lc}
        \hline
        Method & Accuracy \\
        \hline
        \textsc{Scanner} \small{\citep{cheng-etal-2017-learning}} & 82.8\%\\
        \textsc{Coarse2Fine} \small{\citep{dong-lapata-2018-coarse}} & 83.2\%\\
        \hline  
        \textsc{Seq2Seq} \small{\citep{dong-lapata-2016-language}} & 80.9\%\\
        \qquad \textbf{+ PDE} & 84.7\% \\
        \textsc{Seq2Tree} \small{\citep{dong-lapata-2016-language}} & 82.1\%\\
        \qquad \textbf{+ PDE} & \textbf{85.4\%} \\
        \hline
    \end{tabular}
    \caption{Accuracies on \textsc{Formulas}}
    \label{table:formulas_result}
\end{table}
\begin{table}[t]
    \small
    \centering
    \begin{tabular}{lc}
        \hline
        Method & Accuracy \\
        \hline
        \textsc{Seq2Seq} \small{(\citeauthor{dong-lapata-2016-language})} & 47.3\%\\
        \textsc{Seq2Tree} \small{(\citeauthor{dong-lapata-2016-language})} & 49.7\%\\
        \textsc{PointerGenerator} \small{(\citeauthor{see-etal-2017-get})} & 51.0\%\\
        \textsc{Transformer} \small{(\citeauthor{vaswani2017attention})} & 53.4\%\\
        \textsc{Coarse2Fine} \small{(\citeauthor{dong-lapata-2018-coarse})} & 58.1\%\\
        \textsc{HSP} \small{(\citeauthor{zhang-etal-2019-complex})} & 66.2\%\\
        \hline
        \textsc{BaseParser 1 (Seq2Seq+Copy*)} & 58.4\%\\
        \qquad \textbf{+ PDE} & 64.5\% \\
        \hline
        \textsc{BaseParser 2 (Transformer+Copy*)} & 62.8\%\\
        \qquad \textbf{+ PDE} & \textbf{72.2\%} \\
        \hline       
    \end{tabular}
    \caption{Accuracies on \textsc{ComplexWebQuestions}.}
    \label{table:complex_result}
\end{table}
\subsection{Results and Analysis}
\label{section:results_and_analysis}

We denote our framework as \textsc{PDE}~(i.e., Parsing via Divide-and-conquEr) and make a comparison against several previously published systems. We use accuracy as the evaluation metric.
\subsubsection{\textsc{Geo} and \textsc{Formulas}}
Table \ref{table:geo_result} presents the results on \textsc{Geo}.
Compared with previous work using syntax-aware decoders, \textsc{PDE} performs competitively whereas adopts a relatively simple Seq2Seq model. For the Seq2Seq/Seq2Tree base parser, our segmentation mechanism brings accuracy gains of 5.1\% and 4.7\%, respectively.
Table \ref{table:formulas_result} presents results on \textsc{Formulas}, where we observe similar tendencies.
For the Seq2Seq/Seq2Tree base parser, our segmentation mechanism brings accuracy gains of 3.8\% and 3.3\%, respectively.
\subsubsection{\textsc{ComplexWebQuestions}}
Results on \textsc{ComplexWebQuestions} are shown in Table~\ref{table:complex_result}.
We compare \textsc{PDE} against the state-of-the-art model HSP~\citep{zhang-etal-2019-complex} as well as some other baseline models.
The results show that \textsc{PDE} is superior to all baseline models.
For \textsc{BaseParser 1} (\textsc{Seq2Seq + Copy*}), our segmentation mechanism (+ PDE) achieves an accuracy gain of 6.1\% (from 58.4\% to 64.5\%).
For \textsc{BaseParser 2} (\textsc{Transformer + Copy*}), the accuracy gains brought by PDE is 9.4\% (from 62.8\% to 72.2\%).
All these ``+ PDE'' results in Table \ref{table:geo_result}, \ref{table:formulas_result} and \ref{table:complex_result} show that: our framework can consistently boost the performance of different neural semantic parsers in different semantic parsing tasks.

\begin{table}[t]
    \small
    \centering
    \begin{tabular}{lcc}
        \hline
        Method & \textsc{Geo} & \textsc{Formulas} \\
        \hline
        \textsc{Seq2Seq} & 63.1\% & 59.7\%\\
        \qquad + GECA & 68.2\% & 64.6\%\\
        \qquad + PDE (only DA) & 79.1\% & 66.8\%\\
        \qquad \textbf{+ PDE} & \textbf{81.2\%} & \textbf{72.7\%}\\
        \hline
        \textsc{Seq2Tree} & 48.7\%  & 63.8\%\\
        \qquad + GECA & 60.3\%  & 67.5\%\\
        \qquad + PDE (only DA) & 79.4\%  & 74.2\%\\
        \qquad \textbf{+ PDE} & \textbf{80.5\%}  & \textbf{77.6\%} \\
        \hline
    \end{tabular}
    \caption{Evaluation of compositional generalization ability on \textsc{Geo}~(compositional split) and \textsc{Formulas}~(unseen subset).}
    \label{table:compose}
\end{table}

\begin{table}[!t]
    \small
    \centering
    \begin{tabular}{lcc} 
        \hline
        Model & \textsc{Seen (88.1\%)} & \textsc{UnSeen (11.9\%)}\\
        \hline
        HSP & 69.8\% & 40.8\%\\
        \hline
        \textsc{BaseParser 2} & 67.7\% & 27.1\%\\
        \quad\textbf{+ PDE} & \textbf{74.4\%} & \textbf{56.3\%}\\
        \hline  
    \end{tabular}
    \caption{Evaluation of compositional generalization ability on \textsc{ComplexWebQuestions}~(unseen subset).}
    \label{table:complex_unseen}
\end{table}

\begin{table}[t]
    \small
    \centering
    \begin{tabular}{lcccc} 
        \hline
        \multirow{2}*{Model} & \textsc{Conj} & \textsc{Nest} &\textsc{Sup} & \textsc{Compar} \\
        &(41.5\%)&(47.6\%)& (5.0\%)&(6.0\%)\\
        \hline
        \textsc{HSP}  & 66.2\% & 68.3\%& \textbf{68.3\%} & 46.6\%\\
        \hline
        \textsc{BaseParser 2} & 64.8\% & 62.2\% & 53.2\% & 62.2\%\\
        \quad \textbf{+ PDE} & \textbf{71.2\%} & \textbf{75.7\%} & 49.1\% &\textbf{70.3\%}\\
        \hline  
    \end{tabular}
    \caption{Accuracies of different question types on \textsc{ComplexWebQuestions} dataset. }
    \label{table:complex_percent}
\end{table}
\begin{table}[t]
    \small
    \centering
    \begin{tabular}{|l|c|p{5cm}|} 
        \hline
        Label & Count & Examples\\
        \hline
        \multirow{4}{*}{Correct} & \multirow{4}{*}{41} & What other films have \textcolor{green}{\textbf{the actor who played Periwinkle}} been in?\\
        \cline{3-3}
        & & \textcolor{green}{\textbf{What famous facility in Charlotte}} is home to the Carolina Cougar? \\
        \hline
        \multirow{7}{*}{Error} & \multirow{7}{*}{9} & \textcolor{red}{\textbf{Who is the person nominating for an award for cigarettes}} and chocolate milk currently married to?\\
        \cline{3-3}
        & & \textcolor{red}{\textbf{Which movie stars both Mario Lopez}} and Erick Estrada? \\
        \cline{3-3}
        & & \textcolor{red}{\textbf{When did the Red Sox}} win their first pennant? \\
        \hline  
    \end{tabular}
    \caption{Human evaluation of utterance segmentation on a random sample of 50 test questions.}
    \label{table:complex_examples}
\end{table}
\subsubsection{Compositional Generalization Ability}
To demonstrate the compositional generalization ability of \textsc{PDE}, we conduct evaluation on the compositional split for \textsc{Geo} and unseen split for \textsc{ComplexWebQuestions} and \textsc{Formulas}~(see details in section~\ref{section:datasets}). In Table \ref{table:compose}, we show results on \textsc{Geo} and \textsc{Formulas}, together with other two data augmentation baselines. The training data of ``+\textsc{GECA}"~\cite{andreas2019good} 
is augmentated by the protocol which seeks to provide a compositional inductive bias in sequence models. To further validate the effectiveness of PDE, we conduct another baseline ``+only DA", which only trained on the mixture of real data and pseudo supervision~(generated by the PDE).
Compared to those baselines, PDE provides explicit alignments between spans and meaning representations, thus achieving the best performance. For \textsc{ComplexWebQuestions}, we make a comparison with \textsc{HSP}~\cite{zhang-etal-2019-complex} and show the results in Table~\ref{table:complex_unseen}. As we can see, the results on all three datasets consistently prove that PDE brings significant accuracy gains~(\textsc{Geo} $63.1\to 81.2$, \textsc{Formulas} $59.7\to 72.7$, \textsc{ComplexWebQuestions} $27.1\to 56.3$) on splits which require compositional generalization.
\subsubsection{Impact of Different Question Types}
As there are four question types (conjunction, nesting, superlative, and comparative) in \textsc{ComplexWebQuestions}, we further investigate the impact of question types on model performances (Table \ref{table:complex_percent}).
The segmentation mechanism improves the accuracy by a large margin on \textsc{CONJ/NEST/COMPAR} questions, while it is not good at dealing with \textsc{SUP} questions.
\subsubsection{Qualitative Analysis of Segmentation}
To verify whether \textsc{PDE} can provide meaningful segmentations of complex utterances, we conduct a human evaluation on a sample of 50 test questions: given spans predicted by \textsc{PDE}, three non-native, but fluent-English speakers are asked to label whether the segmentation is meaningful or not. 
These questions are all from \textsc{ComplexWebQuestions} dataset.
\textsc{PDE} performs well on segmenting 41 out of 50 questions, which indicates the ``segment-and-parse'' mechanism can explicitly capture some meaningful compositional semantics (Table \ref{table:complex_examples}).
We observe three types of error cases:
\begin{enumerate}
\item \emph{Conjunction Scope.}
For example, in $E_1$ (the first error case in Table \ref{table:complex_examples}), the segmentation model mistakenly regards that the scope of conjunction ``\emph{and}'' is the whole question, while the real scope is ``\emph{cigarettes and chocolate milk}''.
In our future work, we expect to solve this problem through incorporating syntactic information (e.g., syntax tree) into the utterance segmentation model.
\item \emph{Shared Predicates.}
For example, in $E_2$, the reduced question is ``\emph{movie and Erick Estrada?}'', which lacks a predicate (``\emph{stars}'').
In our future work, we expect to solve this problem through measuring completeness of reduced utterances and accordingly performing span substitution.
\item \emph{Superlative Constraint.}
For example, $E_3$: ``\emph{When did the Red Sox win their pennant}'' constrained by a superlative ``\emph{first}'', which cannot be linearly segmented.
To address this problem, we need to guide PDE not to segment such utterances (through predicting $s=x$) and leave them to the base parser.
This is a general principle for dealing with complex language phenomena that exposes the limitation of linear segmentation in complex utterances. We also leave this to future work.
\end{enumerate}
\section{Conclusion}
In this paper, we propose a novel framework for boosting neural semantic parsers via iterative utterance segmentation.
The insight is a bottom-up divide-and-conquer mechanism, which significantly improves the compositional generalization ability and interpretability of neural semantic parsers.
Considering the usual absence of labeled data for utterance segmentation, we propose a cooperative training method to tackle this problem.
Experimental results show that our framework consistently improves the performance of different neural semantic parsers across tasks.

In the future, we plan to improve the robustness of our framework for various complex language phenomena.
We also plan to apply this framework to more semantic parsing tasks such as text-to-SQL and text-to-code.

\bibliography{aaai21}
\end{document}